\documentclass[conference]{IEEEtran}
\IEEEoverridecommandlockouts
\usepackage{cite}
\usepackage{amsmath,amssymb,amsfonts}
\usepackage{algorithmic}
\usepackage{graphicx, multirow, tabularx}
\usepackage{textcomp}
\usepackage{xcolor}
\usepackage[switch]{lineno}
\usepackage{url, hyperref, booktabs}
\usepackage{adjustbox}
\usepackage{float}
\usepackage[T1]{fontenc}
\usepackage[utf8]{inputenc, ragged2e}
\usepackage{tikz}
\usepackage{pifont}
\usepackage{scalerel}
\usepackage{ifthen}
\usepackage{algorithm}
\usepackage{algorithmic}

\newcommand{\etal}{\emph{et al.}}

\def\BibTeX{{\rm B\kern-.05em{\sc i\kern-.025em b}\kern-.08em
    T\kern-.1667em\lower.7ex\hbox{E}\kern-.125emX}}
\begin{document}

\title{FHRFormer: A Self-supervised Transformer Approach for Fetal Heart Rate Inpainting and Forecasting}



\author{Kjersti Engan*$^{1}$, Neel Kanwal*$^{1}$, Anita Yeconia $^{1,2}$, Ladislaus Blacy$^{2}$\\ Yuda Munyaw$^{1,2}$, Estomih Mduma$^{2}$, Hege Ersdal$^{1,3}$\\
$^1$University of Stavanger, Stavanger, Norway\\
$^2$Haydom Lutheran Hospital, Haydom, Tanzania\\
$^3$Stavanger University Hospital, Stavanger, Norway\\
Email: \{kjersti.engan, neel.kanwal\}@uis.no\\
* These authors contributed equally to this work.}

\maketitle
\begin{abstract}

Approximately 10\% of newborns require assistance to initiate breathing at birth, and around 5\% need ventilation support. Fetal heart rate (FHR) monitoring plays a crucial role in assessing fetal well-being during prenatal care, enabling the detection of abnormal patterns and supporting timely obstetric interventions to mitigate fetal risks during labor.  Applying artificial intelligence (AI) methods to analyze large datasets of continuous FHR monitoring episodes with diverse outcomes may offer novel insights into predicting the risk of needing breathing assistance or interventions.
Recent advances in wearable FHR monitors have enabled continuous fetal monitoring without compromising maternal mobility. However, sensor displacement during maternal movement, as well as changes in fetal or maternal position, often lead to signal dropouts, resulting in gaps in the recorded FHR data. Such missing data limits the extraction of meaningful insights and complicates automated (AI-based) analysis.
Traditional approaches to handle missing data, such as simple interpolation techniques, often fail to preserve the spectral characteristics of the signals. In this paper, we propose a masked transformer-based autoencoder approach to reconstruct missing FHR signals by capturing both spatial and frequency components of the data. The proposed method demonstrates robustness across varying durations of missing data and can be used for signal inpainting and forecasting.
The proposed approach can be applied retrospectively to research datasets to support the development of AI-based risk algorithms.  In the future, the proposed method could be integrated into wearable FHR monitoring devices to achieve earlier and more robust risk detection.

\end{abstract}


\section{Introduction}





Despite advances in perinatal care, approximately 10\% of newborns require assistance to initiate breathing at birth, with 5\% needing advanced ventilatory support~\cite{dawes2020, wall2009neonatal}. This persistent clinical challenge highlights the indispensable role of fetal well-being assessment during labor, where continuous fetal heart rate (FHR) monitoring is a pivotal tool for detecting neonatal complications~\cite{urdal2019noise, nageotte2015fetal}. By enabling early detection of hypoxia, acidosis, and other markers of fetal distress, continuous FHR analysis empowers clinicians to intervene proactively, mitigating the risk of neonatal deaths~\cite{nageotte2015fetal, alfirevic2017}. While traditional cardiotocography (CTG) machines effectively detect fetal distress early, their high cost and need for specialized infrastructure make them impractical for many low-resource environments~\cite{mwakawanga2024barriers}. The emergence of wearable FHR monitoring devices, such as the Moyo FHR monitor~\footnote{https://shop.laerdalglobalhealth.com/}, addresses this gap in prenatal care by providing portable, cost-effective solution that enables extended, non-invasive monitoring without substantially limiting maternal mobility 
~\cite{katebijahromi2021detection, alim2023wearable}. 
However, this increased mobility introduces other limitations, most notably signal dropouts caused by sensor displacement during maternal movement. These interruptions can result in missing data within the FHR recordings, which can compromise clinical interpretation and reduce the reliability of artificial intelligence (AI)-based analyses used to predict neonatal outcomes ~\cite{ghosh2024multi, mccoy2025intrapartum}.  

\begin{figure*}[h!]
    \centering
    \includegraphics[width=0.95\textwidth]{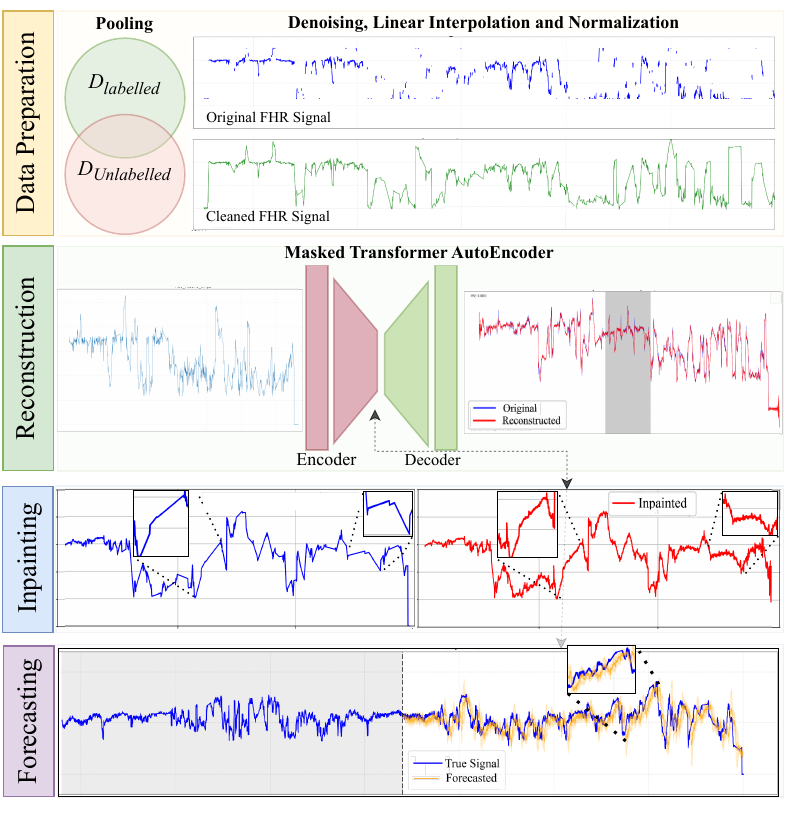}
    \caption{The application of transformer-based autoencoders in this research work: The architecture is trained on the preprocessed version of FHR data, which can perform inpainting and forecasting tasks.
    \emph{\textbf{Data preparation:}} The pooled datasets are preprocessed for Doppler noise and linearly interpolated before performing min-max normalization. \emph{\textbf{Reconstruction:}} A transformer-based encoder-decoder trained in a self-supervised fashion to reconstruct FHR signals by masking different signal areas in each iteration.
    \emph{\textbf{Inpainting:}} Using the trained transformer-based autoencoder to reconstruct the original FHR for linearly interpolated regions to improve signal quality.
    \emph{\textbf{Forecasting:}} Feeding the context of the FHR signal to forecast iteratively for upcoming segments.}
    \label{fig:main}   
\end{figure*}

Missing data adversely impacts both time-frequency and AI-based analysis of FHR signals~\cite{barzideh2018estimation}. This is due to the potential loss of important temporal coherence and spectral features, such as short- and long-term baseline variability and deceleration patterns, that are essential for accurate risk stratification~\cite{barzideh2018estimation, rao2024automatic, cockburn2024clinical}. Therefore, addressing the signal gaps is essential to ensure the reliability of next-generation AI-enabled FHR monitoring systems~\cite{barzideh2018estimation, ghosh2024multi}. Conventional inpainting methods, such as linear or spline interpolation ~\cite{spilka2012using}, ~\cite{spilka2014discriminating}, often oversimplify the physiological complexity of FHR dynamics and fail to preserve critical time-frequency relationships during prolonged signal dropouts, leading to inadequate estimation of features indicative of fetal distress. 
More advanced techniques, including wavelet-based methods~\cite{spyridou2007analysis} and sparse learning algorithms~\cite{barzideh2018estimation}, have demonstrated improved reconstruction accuracy. However, they introduce other challenges, such as performance inefficiency for longer dropouts or fetal state transitions. 

In parallel, AI-based FHR analysis also faces fundamental limitations when relying on conventional architectures such as recurrent neural networks (RNNs) and convolutional neural networks (CNNs)~\cite{xu2023research, tahir2025bridging}. While these models are effective at detecting short-term patterns, they often struggle to capture long-range temporal dependencies that are critical for accurate FHR interpretation. Such shortcomings might compromise inpainting and forecasting performance, especially during prolonged missing data gaps. 

To address these gaps, we propose a self-supervised transformer-based autoencoder framework, \emph{FHRFormer}, tailored for FHR signal reconstruction and forecasting, as illustrated in Figure~\ref{fig:main}. Trained on masked FHR data with focal-frequency loss~\cite{jiang2021focal}, FHRFormer learns both spectral and morphological representations. Unlike sequence-dependent RNNs or locality-constrained CNNs, transformers leverage multi-scale attention mechanisms~\cite{vaswani2017attention, kanwal2022attention} to globally model temporal and spectral dependencies. The frequency-aware loss enables robust reconstruction of missing segments while preserving \emph{time-domain coherence} (e.g., baseline stability, deceleration morphology) and \emph{frequency-domain characteristics} (e.g., variability power spectra). 
Integrated with the Moyo device, this framework enables the acquisition of higher-quality FHR data and introduces intelligent forecasting capabilities for early warning and intervention. 

The main contributions of this paper are: 

\begin{figure*}[ht!]
    \centering
    \includegraphics[width=0.95\linewidth]{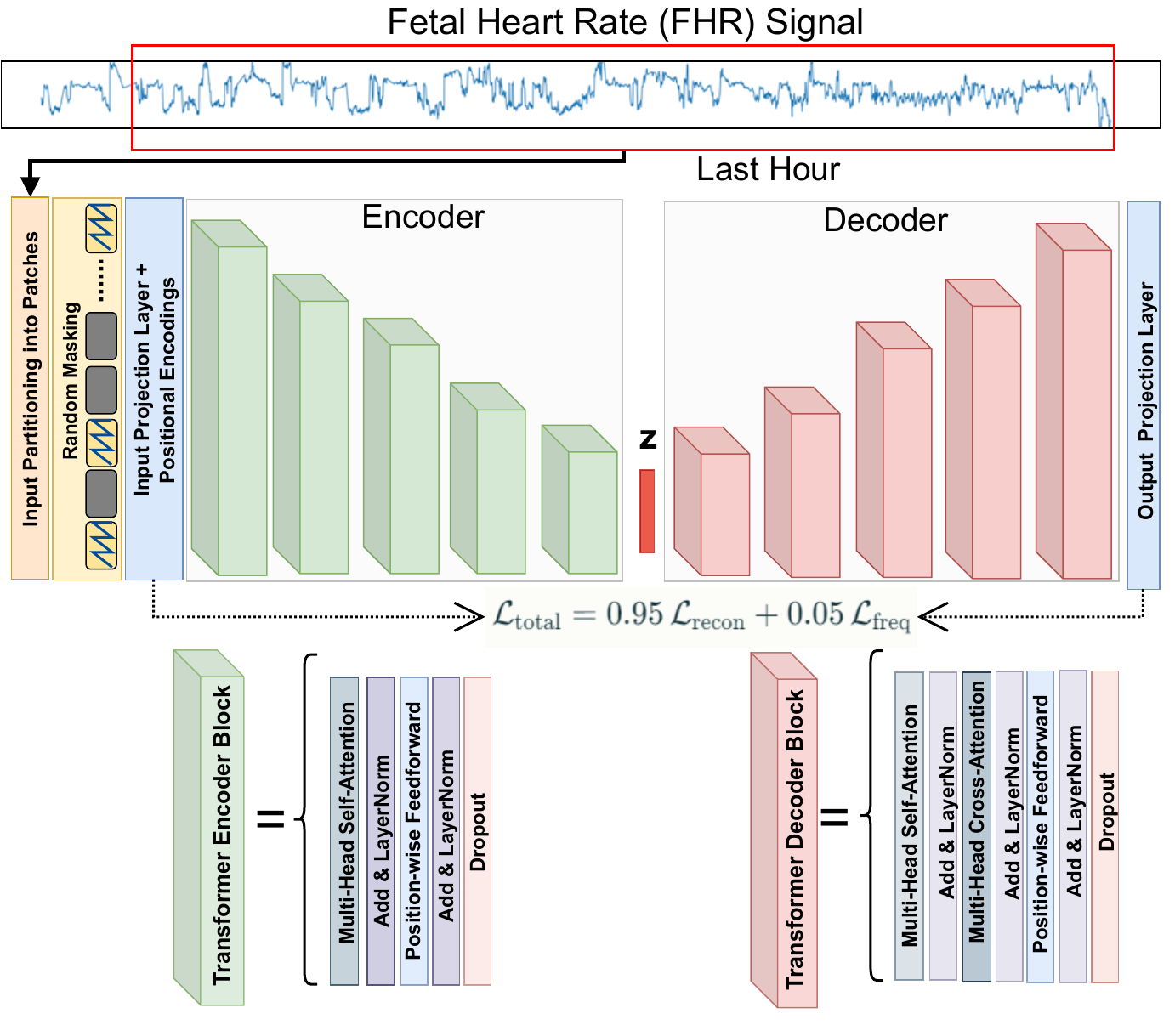}
    \caption{An overview of the 'FHRFormer' architecture. The architecture takes the fetal heart rate signal and divides it into patches. The patches are then embedded and masked before linearly projecting into the encoder. The encoder has five transformer blocks, which produce a compact representation that is then used by the decoder, which contains five transformer blocks. The final reconstruction is compared with the input using a hybrid loss.}
    \label{fig:architecture}
\end{figure*}

\begin{enumerate}
    \item A transformer-based encoder-decoder architecture for reconstructing FHR signals using synthetic masking. 
    \item A self-supervised inpainting strategy to recover missing FHR signal segments. 
    \item A forecasting pipeline to predict FHR evolution,  enhancing utility for AI-based risk assessment. 
\end{enumerate}

 The remainder of this paper is structured as follows. Section~\ref{sec:related} presents recent studies on traditional, early machine learning, and deep learning approaches for FHR analysis. Section~\ref{sec:methodology} provides details on data materials, the preprocessing, methodology, and evaluation metrics. Section~\ref{sec:results} shows experimental results and discusses the applications. Finally, Section~\ref{sec:conclusion} concludes this paper with highlights of some limitations and future work.

\section{Related Work} \label{sec:related}






The problem of handling missing data in FHR monitoring has inspired the development of various techniques, ranging from classical signal processing to cutting-edge deep learning methods. To deliver a clear synthesis, we divide prior work into two categories: i) traditional signal processing and early machine learning approaches, which laid the groundwork for FHR reconstruction, and ii) modern deep learning (DL) approaches, which unlock new possibilities for modeling complex physiological signals.

\subsection{Signal Processing and Early Machine Learning Approaches}

Earlier works focusing on the missing data challenge in FHR signals relied heavily on signal processing techniques, often prioritizing simplicity over physiological fidelity. Linear interpolation~\cite{spilka2012using}, for instance, was a common choice due to its ease of implementation. However, it is a naive approach, as it does not capture the intricate variability of FHR dynamics, such as baseline fluctuations or transient decelerations~\cite{ribeiro2021non, campos2024fetal}. While more sophisticated interpolation methods (e.g., spline or cubic interpolation~\cite{cesarelli2011psd}) can better preserve signal smoothness and reduce spectral distortion compared to linear methods, they remain prone to oversmoothing transient patterns like abrupt decelerations or oscillations~\cite{campos2024fetal}. This oversimplification often makes interpolated signals unreliable, as critical spectral features are lost. In short, interpolation-based methods have low computational cost but are effective for smooth signals and smaller missing gaps. Lately, wavelet-based methods~\cite{spyridou2007analysis} emerged as a more sophisticated alternative, decomposing signals into time-frequency components to better preserve localized patterns. While effective for short gaps, wavelet-based approaches also struggled with longer dropouts, where the trade-off between time and frequency resolution became pronounced, particularly in nonstationary FHR signals driven by fetal behavioral changes~\cite{zhu2021retracted}.

To overcome these limitations, sparse representation techniques emerged as a promising direction. Oikonomou \etal~\cite{oikonomou2013adaptive} proposed a two-step adaptive method for reconstructing missing FHR samples using an empirical dictionary. Their method assumed a local signal stationarity, which may not hold during an active labor phase. Poian \etal~\cite{da2015sparse} introduced a dictionary learning method for accurate FHR estimation using sparse decomposition of Gaussian-like functions. Barzideh \etal~\cite{barzideh2018estimation} explored shift-invariant dictionary learning, constructing adaptive bases to reconstruct FHR signals by capturing variable shifts and transient patterns. However, their dependence on fixed dictionary sizes posed challenges when faced with variable dropout durations, which are common in real-world monitoring scenarios. In short, sparse learning methods have a higher computational cost and are error-prone to fetal state transitions.

The advent of machine learning in biomedical signal analysis brought new methods for handling missing FHR data, with early efforts focusing on manual feature selection. Georgieva \etal~\cite{georgieva2013artificial} performed principal component analysis to reduce feature space and trained a feed-forward artificial neural network for classifying birth asphyxia. Zhao \etal~\cite{jcm7080223} performed comprehensive feature selection to train a decision tree, support vector machine, and adaptive boosting algorithm for fetal asphyxia classification. Dash \etal~\cite{dash2014fetal} proposed a probabilistic method using Bayes' rule to classify FHR features. These foundational works reveal the critical need for architectures that can automatically capture hierarchical FHR patterns, paving the way for decoding FHR's physiological complexity.

\subsection {Deep Learning Approaches}
Recent advances in deep learning have reshaped the landscape for biomedical time-series analysis, offering tools better suited to the challenges of FHR monitoring. CNNs were widely adopted for feature extraction and classification, particularly for baseline determination and anomaly detection. For instance, Lin \etal~\cite{lin2024deep} proposed LARA (Long-term Antepartum Risk Analysis) system, which combined CNNs with information fusion operators to interpret FHR through attention visualization and deep feature analysis. Similarly, Zhong \etal~\cite{zhong2022ctgnet} proposed CTGNet, which utilized CNNs for FHR baseline calculation and acceleration/deceleration detection, leveraging preprocessing techniques to filter artifacts and classify signal segments. However, it is well known that CNNs often struggle with modeling long-range dependencies, limiting their ability to capture temporal patterns across extended FHR recordings~\cite{lin2024deep, zhao2019deepfhr}. Recurrent architectures, including long short-term memory (LSTM) networks, resolve this challenge of temporal patterns to some extent by leveraging their ability to model short-term dependencies in FHR signals~\cite{petrozziello2018deep}. Boudet \etal~\cite{boudet2022use}, FSDROP, a Gated Recurrent Unit (GRU)-based model to detect false signals in FHR, showed promising results for some cases. Since RNNs are constrained by fixed-length context and sequential processing, they are inadequate for parallel processing, hindering their usability for real-time FHR processing. 

Transformer-based methods have gained prominence for their ability to model multi-scale dependencies through attention mechanisms~\cite{vaswani2017attention}. 
In broader medical time-series domains, like electrocardiography (ECG) and electroencephalography (EEG), transformers~\cite{zhang2022maefe,zhang2023self, wang2024convolutional} have shown promising results for various tasks by attending to both local and global patterns. However, their potential in fetal monitoring remains largely unexplored, especially for inpainting missing segments and forecasting future trends.
While self-supervised learning, particularly masked autoencoding, offers a powerful approach for leveraging unlabeled data to learn robust representations, the FHR-specific applications of self-supervised transformers are currently lacking. This absence leaves a significant unmet need for effectively handling signal dropouts commonly encountered in FHR data from wearable devices.

Our work bridges these gaps by proposing \emph{FHRFormer}, a masked-transformer architecture tailored for FHR inpainting and forecasting. The proposed method overcomes limitations of prior traditional approaches by integrating frequency-aware loss functions~\cite{jiang2021focal} and self-supervised pretraining on unlabeled FHR data. This dual focus on temporal coherence and spectral fidelity positions our framework as a transformative tool for real-time fetal monitoring, particularly in addressing the nonstationary challenges.


\section{Data Materials and Methodology}  \label{sec:methodology}

This section presents details on the FHR data used for training and validation, describes data preprocessing for pre-training, the structure of the transformer-based autoencoder, fine-tuning for the forecasting task, the experimental setup, and the evaluation metrics used for performance evaluation. 

\subsection{Data Material}
The FHR data were obtained through the Safer Births project~\footnote{saferbirths.com}, a collaborative research initiative involving international research institutions and hospitals in Tanzania. Data collection occurred in two phases at two urban and one rural Tanzanian hospital between October 2015 to June 2018, and December 2019 to July 2021, encompassing a total of 5,225 recorded labors. 
This study focuses exclusively on labors initially assessed as normal upon hospital admission. The data was acquired using the Laerdal Moyo FHR monitor~\cite{laerdal_moyo}. This device comprises a primary unit displaying the measured heart rate to healthcare personnel and a sensor unit integrating a Doppler ultrasound sensor and an accelerometer. The sensor unit is secured to the mother using an elastic strap. The device is programmed to alert healthcare personnel if the detected FHR remains outside the range of 110-160 beats per minute for 10 minutes or outside the range of 100-180 beats per minute for 3 minutes. The FHR is measured via a $1\,\mathrm{MHz}$, $5\,\mathrm{mW/cm^2}$ pulsed wave Doppler ultrasound sensor, sampled at $2\,\mathrm{Hz}$, generating a discrete FHR signal, $\mathbf{x} \in \mathbb{R}^{L \times 1}$  where $L$ is the FHR signal length. 

\subsection{Preprocessing}
Doppler shift error can result in a doubling or halving of the perceived frequency, and thereby the estimated heart rate.  
Additionally, periods of missing data may occur due to sensor displacement caused by maternal movement, as previously discussed. To mitigate these issues, the 
noise reduction algorithm by Urdal \etal~\cite{urdal2019noise} was employed during preprocessing. This method addresses frequency-doubling/halving artifacts and applies linear interpolation as an initial step to fill missing signal segments. A binary mask indicating the locations of missing data was retained to enable refined inpainting using the proposed FHRFormer framework.

All FHR signals were trimmed or zero-padded to a fixed length of 7,200 time steps, corresponding to one hour of physiological time series data. This standardization focuses on the final hour preceding birth, which is considered the most clinically relevant period.

The preprocessed FHR dataset, $\mathcal{X}=\{{\mathbf x}^{(j)}\}$, where $j$ is the episode index, often omitted for simplicity, was divided into three non-overlapping subsets at the episode (patient) level, subsets: Training set $\mathcal{X}_{tr}$, a validation set $\mathcal{X}_{val}$, and a test set $\mathcal{X}_{test}$, consisting of 4486, 369, and 370 FHR episodes, respectively.


\subsection{FHRFormer Architecture}
\label{sec:FHRFormer}
Figure~\ref{fig:architecture} shows a detailed overview of the proposed FHRFormer architecture along with the training strategy. The following sections further describe the components of the architecture.

\subsubsection{Input Patchification}
Given a univariate FHR timeseries signal $\mathbf{x} \in \mathbb{R}^{L \times d}$, where $L$ is the FHR length and $d$ is the signal dimensionality (here $d=1$). The input is first divided into $N=L/p$ non-overlapping segments, or patches, of length $p_s$.

\begin{equation}
\mathbf{x} = [\mathbf{x}_1, \mathbf{x}_2, \ldots, \mathbf{x}_N], \quad \mathbf{x}_i \in \mathbb{R}^{p_s \times d}
\end{equation}

The purpose of patching is to allow the transformer model to process manageable local segments while still retaining the ability to learn dependencies across the entire time sequence. We have experimented with multiple patch sizes, $p_s =(30, 60, 120, 240, 480$). 

\begin{table}[ht!]
    \centering
    \caption{The configuration and hyperparameters used for the training of the proposed FHRFormer model.}
     \resizebox{0.49\textwidth}{!}{
    \begin{tabular}{||l| l||}
    \hline
       {Parameter} & Value \\ 
       \hline
        Number of encoder layers & 5 \\
        Number of decoder layers & 5 \\
        Number of attention heads & 16 \\
        Dimension of intermediate layer & 1024 \\ 
        Features in decoder input ($d_{model}$) & 512 \\ 
        Input patch size ($p_s$) & \{30, 60, 120, 240, 480\} \\
        Total length of input ($L$) & 7200 \\
        Masking ratio ($\gamma$ \%) & \{5, 10, 15, 20, 25, 30, 35\} \\
        Input normalization & min-max \\
        Early stopping (patience $Pat$) & 20 epochs\\ 
        Batch size & 128 \\
        Scheduler & ReduceLRonPlateau with patience=5\\
        Optimizer & Adam \\
        Learning rate ($\eta$) & 0.0001 \\
        Dropout regularization & 0.1 \\
        Loss function ($\mathcal{L}$) & 0.95$\times$($\mathcal{L}_{recon}$) + 0.05$\times$ ($\mathcal{L}_{freq}$)\\
        \hline
    \end{tabular}}
    \label{tab:implementation}
\end{table}

\subsubsection{Randomized Masking Strategy}
The masking strategy promotes robust feature learning in a self-supervised fashion by encouraging the model to infer and reconstruct missing information from the surrounding context. 


Masking is performed randomly at the patch level, using a binary mask vector $\mathbf{m} \in \{0, 1\}^N$, where $m_i = 0$ indicates that patch i is masked.
We define the sets:
$\mathcal{M} = \{i \mid m_i = 0\}$ as the indices of masked patches
$\mathcal{U} = \{i \mid m_i = 1\}$as the indices of unmasked patches. 
To ensure sufficient learning signal, at least one patch is always masked. The parameter $\gamma$ controls the target masking ratio, such that approximately $\gamma \cdot N$ patches are masked in each input signal.

\subsubsection{Input Embedding and Positional encoding}
Each patch $\mathbf{x}_i$ is first flattened and then projection into a higher-dimensional latent space using a learnable linear projection.

\begin{equation}
\mathbf{e}_i = \mathbf{W}_{in} \cdot \mathrm{vec}(\mathbf{x}_i) + \mathbf{b}_{in}
\end{equation}

where $\mathbf{W}_{in} \in \mathbb{R}^{d_{\text{model}} \times (p_s \cdot d)}$ are learnable input projection weights and $\mathbf{b}_{in}$ is bias, and $d_{\text{model}}$ is the transformer’s feature dimension, i.e. $\mathbf{e}_i \in \mathbb{R}^{d_{\text{model}}}$, via the learnable projection. These projections enable the transformer to represent local signal structure in a rich, learnable latent space. 
To retain positional information, positional  encodings $\mathbf{p}_i$ are added to both masked and unmasked embedded patches. The unmasked patches with positional encoding gives the encoder input:
\begin{equation}
    \tilde{\mathbf{e}}_i = \mathbf{e}_i + \mathbf{p}_i, \; i \in \mathcal{U}
\end{equation}
A single learnable vector $\mathbf{e}_{mask} \in \mathbb{R}^{d_{\text{model}}}$ is shared across all masked patches, used as a placeholder embedding, with $\mathbf{e}_{mask} + \mathbf{p}_i , \; i \in \mathcal{M}$ as input to the decoder only.  
\subsubsection{Transformer Encoder}
The unmasked embedded patches form the input to a stack of transformer encoder blocks. Each encoder block consists of i) Multi-head Self-attention (MHSA), which computes contextualized representations allowing each patch to attend to every other visible patch; ii) Position-wise Feedforward Networks (FFN), which enhance the model’s capacity for nonlinear feature transformation; and iii) residual connections and layer normalization (LN), referred to as Add \& LayerNorm in Figure~\ref{fig:architecture}, are applied following each sub-block for training stability and adding the input to the attention output.
Let $\mathbf{h}_i^{(0)} =  \tilde{\mathbf{e}}_i$.  Formally, for the encoder layer $\ell$


\begin{equation}
\begin{aligned}
{\mathbf{h}}_i^{(\ell,1)} &= \mathrm{LN}\left(\mathbf{h}_i^{(\ell)} + \mathrm{MHSA}\left(\mathbf{h}_i^{(\ell)}\right)\right) \\
\mathbf{h}_i^{(\ell,2)} &= \mathrm{LN}\left({\mathbf{h}}_i^{(\ell,1)} + \mathrm{FFN}\left({\mathbf{h}}_i^{(\ell,1)}\right)\right)
\end{aligned}
\end{equation}

The dropout is applied for stochastic regularization and discourages overfitting by preventing co-adaptation of neurons and encourages the network to learn redundant representations. 
The output of the final encoder layer for patch $\mathbf{x}_{i}$ is denoted as $\mathbf{z}_{i}$, $\mathbf{Z}= [\mathbf{z}_{0}, \mathbf{z}_{i} \ldots ], \; i\in\mathcal{U}$. \\



\subsubsection{Transformer Decoder}



The Transformer decoder receives both the visible encoder outputs and learned mask tokens, and produces embeddings for all patches. However, only the masked patches are reconstructed and contribute to the training objective

The decoder receives the following input: 
\begin{equation}
\mathbf{d}_{i}^{0} = 
\begin{cases}
\mathbf{z}_i + \mathbf{p}i, & i \in \mathcal{U} \\
\mathbf{e}_{mask} + \mathbf{p}_i, & i \in \mathcal{M}
\end{cases}
\end{equation}
with the notation:  $\mathbf{D}^{(0)} = [\mathbf{d}^{(0)}_1, \dots, \mathbf{d}^{(0)}_N].$

The decoder consists of a stack of identical decoder blocks. Each block includes three main sub-layers, with residual connections, layer normalization, and dropout. Given the non-causal nature of the reconstruction task, each embedding can attend to all positions in the sequence, allowing the decoder to reason globally when reconstructing masked regions. 

The $\ell$-th decoder layer is defined through some steps, starting with a self-attention layer: 

\begin{equation}
\mathbf{D}^{(\ell,1)} = \text{LN}\left( \mathbf{D}^{(\ell-1)} +  \text{MHSA}(\mathbf{D}^{(\ell-1)}) \right) 
\end{equation}

Next, the Multi-Head Cross-Attention (MHCA) sub-layer allows each decoded patch to attend to the encoder outputs of the unmasked inputs, $\mathbf{Z}$. The queries are from the output of the MHSA layer, while keys and values are the encoder's representations $\mathbf{Z}$. This enables the decoder to consult both the visible context (from the encoder) and the reconstructed context (from the decoder) at each step:

\begin{equation}
\mathbf{D}^{(\ell,2)} = \text{LN}\left( \mathbf{D}^{(\ell,1)} + \text{MHCA}(\mathbf{D}^{(\ell,1)}, \mathbf{Z}) \right)
\end{equation}


\begin{algorithm}[h!]
\caption{Training of FHRFormer with Early Stopping} \label{alg:MTA}
\begin{algorithmic}[1]
    \STATE \textbf{Input:} Training data 
    $\mathcal{X}_{tr}$, Validation data $\mathcal{X}_{val}$, 
    masking ratio $\gamma$, patch size $p_s$, model parameters $\theta$, learning rate $\eta$, max epochs $E$, patience $Pat{=20}$
    \STATE Initialize optimizer with $\eta$
    \STATE Initialize best validation loss $L_{best} \leftarrow \infty$
    \STATE Initialize epochs since improvement $c \leftarrow 0$
    \FOR{epoch $= 1$ to $E$}
        \STATE Set the model to training mode.
        \FOR{each batch $\mathbf{x}_{b}$ in $\mathcal{X}_{tr}$ } 
            \STATE Normalize $\mathbf{x}_b$
            \STATE Divide $\mathbf{x}_b$ into patches of size $p_s$
            \STATE Mask a random fraction $\gamma$ of each patch; select visible set $V$ and masked set $M$
            \STATE Embed visible patches (V), add positional encoding
            \STATE $\mathbf{z} \leftarrow \text{Encoder}(\{x_i : i \in V\})$
            \STATE Construct decoder input from $\mathbf{z}$ and mask tokens, with positional encoding.
            \STATE $\hat{\mathbf{x}} \leftarrow \text{Decoder}(\mathbf{z}, \{\text{mask tokens}\})$
            \STATE Compute recons. loss: $\mathcal{L}_{\mathrm{recon}}$ as defined in \eqref{eq:rec}
            \STATE Compute frequency loss $\mathcal{L}_{\mathrm{freq}}$ as defined in \eqref{eq:freq}
            \STATE Compute total loss: $\mathcal{L}_{\mathrm{total}}$ as defined in \eqref{eq:tot}
            \STATE Update $\theta$ as $\theta_{best}$ using optimizer 
        \ENDFOR
        \STATE Set model to evaluation mode.
        \STATE Compute validation loss $\mathcal{L}_{\mathrm{val}}$ by averaging over all batches in $\mathbf{x_{val}}$ (repeat steps 4--13)
        \IF{$\mathcal{L}_{\mathrm{val}} < L_{best}$}
            \STATE $L_{best} \leftarrow \mathcal{L}_{\mathrm{val}}$
            \STATE Save current model weights as best model.
            \STATE $c \leftarrow 0$ \COMMENT{Reset patience counter}
        \ELSE
            \STATE $c \leftarrow c + 1$
        \ENDIF
        \IF{$c \geq Pat$}
            \STATE \textbf{Break training loop} \COMMENT{Early stopping triggered}
        \ENDIF
    \ENDFOR
    \STATE \textbf{Output:} Best model parameters $\theta_{best}$
\end{algorithmic}
\end{algorithm}

\begin{table*}[ht!]
    \centering
    \caption{The validation and test results. The performance of the transformer-based autoencoder for different input patch sizes. The best results are highlighted in bold, and the second-best are underlined for each subset.}
    \renewcommand{\arraystretch}{2} 
    \begin{tabularx}{\textwidth}{|>{\Centering\arraybackslash}m{0.5cm} |>{\Centering\arraybackslash}m{1.3cm} |>{\Centering\arraybackslash}m{1.5cm} |>{\Centering\arraybackslash}m{1.5cm} |>{\Centering\arraybackslash}m{1.5cm} |>{\Centering\arraybackslash}m{1.5cm} |>{\Centering\arraybackslash}m{1.5cm} |>{\Centering\arraybackslash}m{1.5cm} |>{\Centering\arraybackslash}m{1.5cm} |>{\Centering\arraybackslash}m{1.5cm} |}
       \cline{1-10}
       \hline\hline
         {} & \textbf{\shortstack{Input\\ Patch\\ Size ($p_s$)}} & \textbf{\shortstack{RL ($\downarrow$)}} & \textbf{\shortstack{PSNR ($\uparrow$)}} & \textbf{\shortstack{SSIM ($\uparrow$)}} & \textbf{\shortstack{FID ($\downarrow$)}} & \textbf{\shortstack{MSE ($\downarrow$)}} & \textbf{\shortstack{RMSE ($\downarrow$)}} & \textbf{\shortstack{MAE ($\downarrow$)}} & \textbf{\shortstack{CC ($\uparrow$)}} \\ 
    \cline{1-10}
    \hline\hline
        \multirow{5}{*}{\rotatebox[origin=c]{90}{\textbf{Validation}}} 
        & 30  & \textbf{0.051}  & \textbf{41.55} & \textbf{0.9981} & \textbf{0.503} & \textbf{0.000071} & \textbf{0.0084} & \textbf{0.0041} & \textbf{0.9992} \\ 
        \cline{2-10}
        & 60  & \underline{0.205}  & \underline{35.63} & \underline{0.9925} & \underline{1.965} & \underline{0.000273} & 0.1653 & \underline{0.0088} & \underline{0.9971} \\ 
        \cline{2-10}
        & 120 & 0.489  & 31.58 & 0.9822 & 4.985 & 0.000694 & \underline{0.0264} & 0.0149 & 0.9925 \\ 
        \cline{2-10}
        & 240 & 1.264 & 27.72 & 0.9615 & 12.051 & 0.001695 & 0.0411 & 0.0244 & 0.9816 \\ 
        \cline{2-10}
        & 480 & 3.232 & 23.01 & 0.9092 & 34.242 & 0.004994 & 0.0707 & 0.0433 & 0.9454 \\ 
        \cline{1-10}
       \hline\hline
        \multirow{5}{*}{\rotatebox[origin=c]{90}{\textbf{Test}}} 
        & 30  & \textbf{0.048}  & \textbf{41.38} & \textbf{0.9984} & \textbf{0.522} & \textbf{0.000072} & \textbf{0.0085} & \textbf{0.0040} & \textbf{0.9993} \\ 
        \cline{2-10}
        & 60  & \underline{0.204}  & \underline{35.56} & \underline{0.9941} & \underline{1.999} & \underline{0.000278} & \underline{0.0166} & \underline{0.0086} & \underline{0.9976} \\ 
        \cline{2-10}
        & 120 & 0.537  & 31.57 & 0.9862 & 4.987 & 0.000952 & 0.02637 & 0.0145 & 0.9942 \\ 
        \cline{2-10}
        & 240 & 1.228 & 27.71 & 0.9703 & 12.041 & 0.001692 & 0.0411 & 0.0240 & 0.9854 \\ 
        \cline{2-10}
        & 480 & 3.262 & 22.95 & 0.9275 & 34.663 & 0.005061 & 0.0711& 0.0435 & 0.9553 \\ 
        \cline{1-10}
        \hline\hline
    \end{tabularx}
    \label{tab1}
\end{table*}

Finally, Position-wise FFN provides depth and non-linearity at each position.
\begin{equation}
\mathbf{D}^{(\ell,3)} = \text{LN}\left( \mathbf{D}^{(\ell,2)} + \left( \text{FFN}(\mathbf{D}^{(\ell,2)}) \right) \right)
\end{equation}


The output of the final decoder layer is denoted as $\mathbf{D} \in \mathbb{R}^{N \times d_{\text{model}}}$, representing contextualized embeddings for all patches (masked and unmasked).

To reconstruct the original FHR signal, a linear output projection layer maps these high-dimensional embeddings back to patch space:
\begin{equation}
    \hat{\mathbf{x}}_i = \mathbf{W}_{\text{out}} \mathbf{d}_i + \mathbf{b}_{\text{out}}, \quad \text{where} \quad \mathbf{W}_{\text{out}} \in \mathbb{R}^{(p_s \cdot d) \times d_{\text{model}}}
\end{equation}
Here, $p_s$ is the patch size and $d$ is the feature dimension of the original signal, thus $\hat{\mathbf{x}}_i \in \mathbb{R}^{p_s \times d}$ is the reconstruction of the $i$-th patch.
Normally, only the masked part is reconstructed and the unmasked part is kept as the original, giving the reconstructed signal as $X^{R} =[x_{0}^{R} \ldots x_{N}^{R}]$ with:  
\begin{equation}
\mathbf{x}_{i}^{R} =
\begin{cases}
\mathbf{x}_i & i \in \mathcal{U} \\
\hat{\mathbf{x}}_{i} , & i \in \mathcal{M}
\end{cases}
\end{equation}

\begin{figure*}[h!]
    \includegraphics[width=0.95\linewidth]{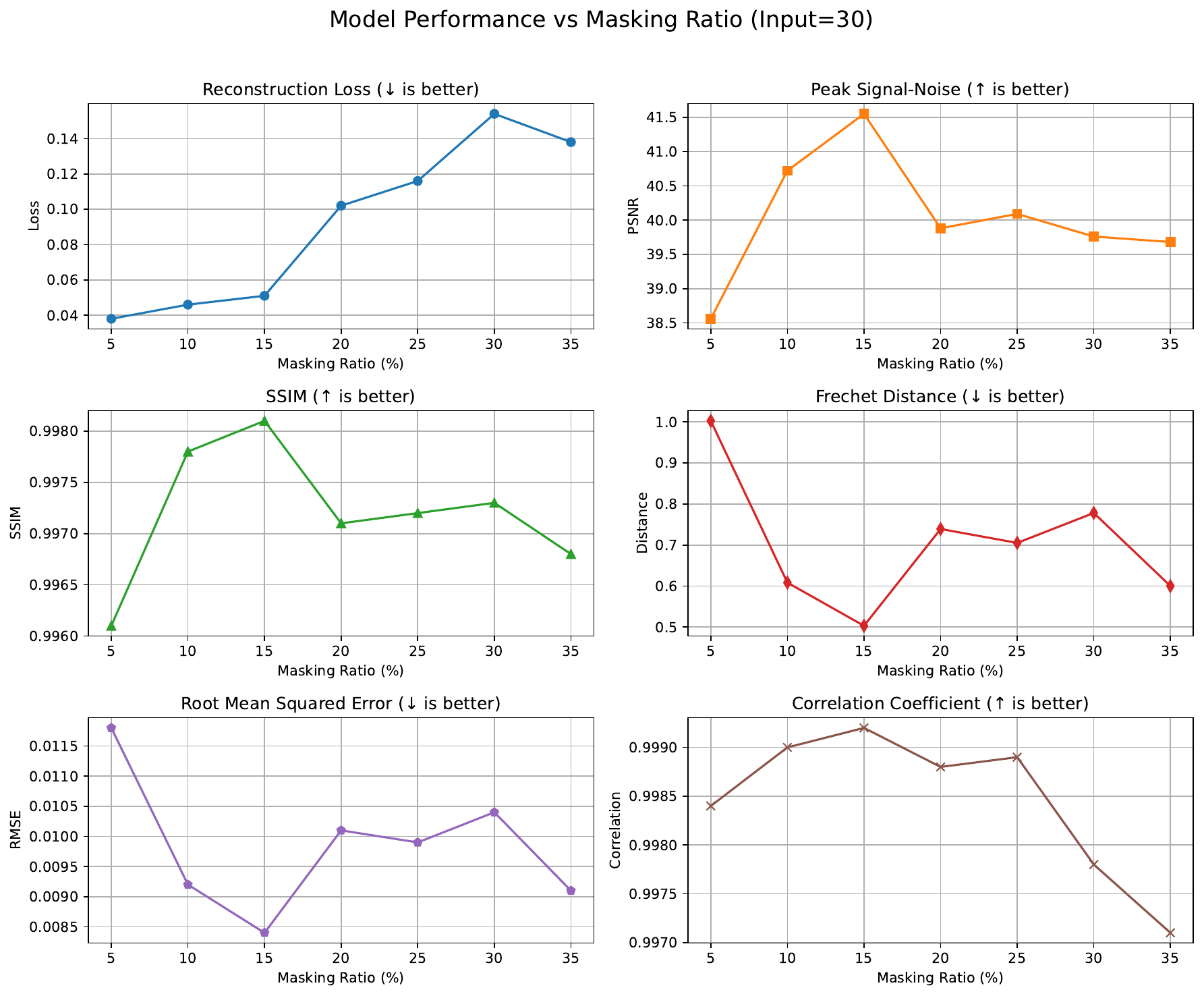}
    \caption{Model performance trends across varying masking ratios (x-axis) with fixed input size 30. The six subplots display Reconstruction Loss (RL), Peak Signal-to-Noise Ratio (PSNR), SSIM, Frechet Distance (FID), Root Mean Squared Error (RMSE), and Correlation Coefficient over y-axes. Each subplot illustrates how the metric changes with masking ratio ($\gamma$), highlighting the trade-offs in model accuracy and quality.}
    \label{fig:masking}
\end{figure*}

\subsubsection{Reconstruction Objective} 
\label{sec:objective}
The objective of FHRFormer is to be able to reconstruct masked regions of the FHR signal as closely as possible. Let $\mathcal{M}$ be the set of indices corresponding to withheld (masked) patches. The \textit{primary loss} is the mean squared error on the masked patches.

\begin{equation} \label{eq:rec}
    \mathcal{L}_{\text{recon}} = \frac{1}{|\mathcal{M}|} \sum_{i \in \mathcal{M}} \|\mathbf{\hat{x}}_i - \mathbf{x}_i\|^2
\end{equation}

To further guide the model toward preserving essential spectral (frequency) physiological characteristics of FHR signals, a frequency-domain loss, namely focal frequency loss~\cite{jiang2021focal}, is combined with the reconstruction loss to develop a hybrid loss function.

\begin{equation} \label{eq:freq}
\mathcal{L}_{\text{freq}} = \frac{1}{N} \sum_{k} \left(1 - e^{-\left||\mathcal{F}(\mathbf{x})_k| - |\mathcal{F}(\mathbf{\hat{x}})_k|\right|}\right)^\beta \cdot \left||\mathcal{F}(\mathbf{\hat{x}})_k| - |\mathcal{F}(\mathbf{x})_k|\right|
\end{equation}

$\mathcal{L}_{\text{freq}}$ is averaged over $N$ frequency components. Each component $k$ contributes to the sum, where $|\mathcal{F}(\mathbf{x})_k|$ and $|\mathcal{F}(\mathbf{\hat{x}})_k|$ denote the amplitude (magnitude) of the $k$-th Fourier transform coefficient for the target signal $\mathbf{x_i}$ and the predicted signal $\mathbf{\hat{x_i}}$, respectively. The parameter $\beta$ (fixed to 1) controls the weighting of the loss based on the discrepancy between the target and predicted amplitudes.

\begin{equation} \label{eq:tot}
    \mathcal{L}_{\text{total}} = \alpha \cdot \mathcal{L}_{\text{recon}} + (1 - \alpha) \cdot \mathcal{L}_{\text{freq}}
\end{equation}

where $0 < \alpha < 1$ (e.g., $\alpha = 0.95$ in our case). The frequency loss encourages the model not only to minimize point-wise errors but also to preserve underlying patterns and periodicity critical to physiological interpretation. The adopted hybrid loss leads to reconstructions that are more realistic in the spectral sense. Algorithm~\ref{alg:MTA} represents the training process for FHRFormer.

\subsection{Implementation Details}

We used Python and PyTorch for scripting and architecture design. Since the architecture was custom-designed, we initialized with random weights. Hyperparameters were explored through a grid search for improved validation metrics. The final chosen parameters were the Adam optimizer with a weight decay of 0.01, the \emph{ReduceRLOnPlateau} scheduler with a learning rate of 0.0001, early stopping of 20 epochs over the validation loss to avoid overfitting, the batch size of 128, and a fixed random seed for reproducibility. Finally, we used the best-performing model weights to report evaluation metrics on the validation and hold-out sets. All chosen hyperparameters for the FHRFormer architecture and training are summarized in Table~\ref{tab:implementation}. All training and inference experiments were performed on a Tesla V100-PCIE with 32 GB.
The source code and model weights are available at \href{https://www.overleaf.com/project/67b84922596fb5f3c7df6aaa}{GitHub}.

\subsection{Evaluation Metrics}
For comprehensive performance assessment of FHRFormer, we report the following metrics: Reconstruction Loss, Peak Signal-to-Noise Ratio (PSNR), Structural Similarity Index Measure (SSIM), Fréchet Inception Distance (FID), Mean Squared Error (MSE), Root Mean Squared Error (RMSE), Mean Absolute Error (MAE), and Correlation Coefficient (CC). 

\textit{Reconstruction Loss (RL)} quantifies the difference between the original and reconstructed FHR, as defined by the loss function in~\eqref{eq:rec}. \textit{Mean Squared Error (MSE)} is calculated as the average squared difference between the target and reconstructed signals, given by $ \mathrm{MSE} = \frac{1}{N} \sum_{i=1}^N (\mathbf{x}_i - \hat{\mathbf{x}}_i)^2 $, where $\mathbf{x}_i$ and $\hat{\mathbf{x}}_i$ denote original and reconstructed values, respectively.

\begin{figure*}[ht!]
    \centering
    \includegraphics[width=0.98\linewidth]{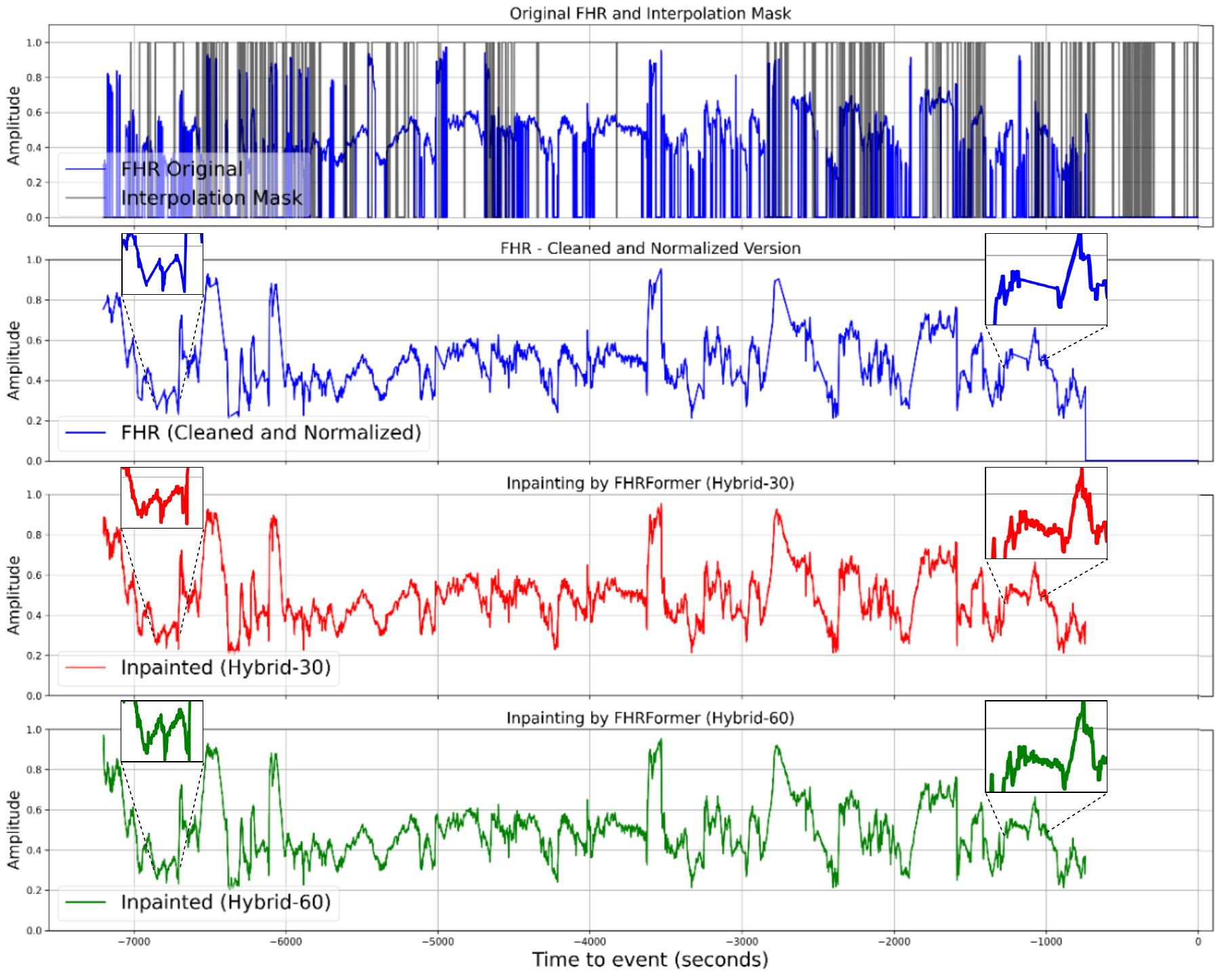}
    \caption{Inpainting application using the  FHRFormer. The first row shows the original FHR signal acquired from the Moyo device. The second row shows the noise-removed and the interpolated version. The third and fourth rows show inpainting performance by the best (Hybrid-30) and the second-best (Hybrid-60).}
    \label{fig:inpainting}
\end{figure*}

\begin{figure*}[h!]
    \centering
    \includegraphics[width=0.98\linewidth]{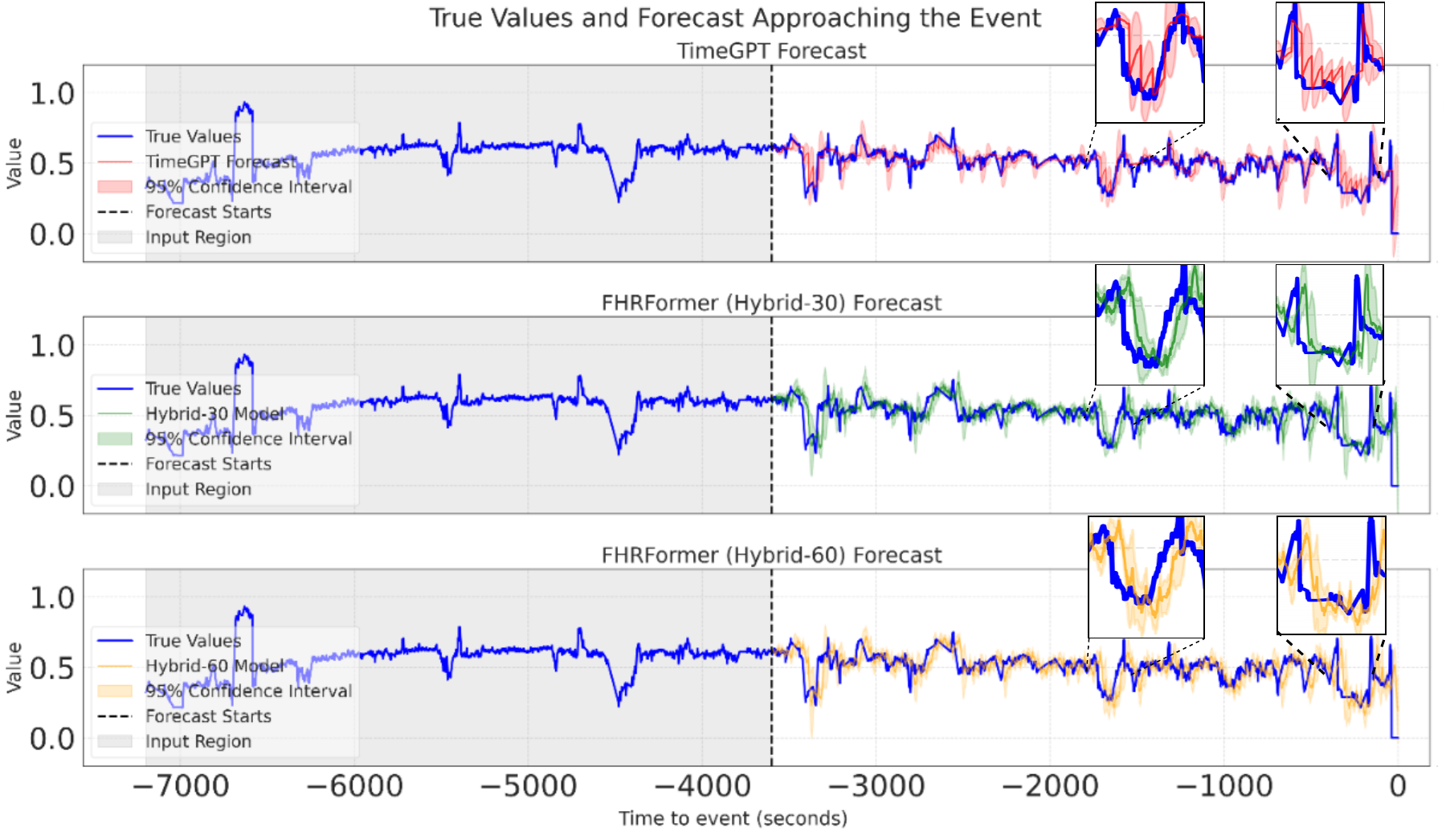}
    \caption{Forecasting application using the FHRFormer. The FHRFormer is fed with a context window (gray region), which is 3600 timesteps (30 minutes of FHR data) as the past horizon, and starts forecasting 30 timesteps (15 seconds of FHR data) in a progressive forecasting style. The first row shows forecasting performance by the TimeGPT~\cite{garza2023timegpt} model. The second row shows forecasting performance by the best FHRFormer, and the last row shows forecasting performance by the second-best FHRFormer.}
    \label{fig:forecasting}
\end{figure*}

\textit{Root Mean Squared Error (RMSE)} provides an interpretable measure by taking the square root of MSE, i.e., $ \mathrm{RMSE} = \sqrt{\mathrm{MSE}} $, allowing direct comparison in the scale of input values. \textit{Mean Absolute Error (MAE)} reflects the average magnitude of reconstruction errors without considering their direction: $ \mathrm{MAE} = \frac{1}{N} \sum_{i=1}^N |\mathbf{x}_i - \hat{\mathbf{x}}_i| $. \textit{Peak Signal-to-Noise Ratio (PSNR)} expresses the reconstruction quality as the ratio between the maximum possible signal and the error power, measured in decibels (dB): $ \mathrm{PSNR} = 10 \cdot \log_{10}\left(\frac{{\max(x)^2}}{\mathrm{MSE}}\right) $. Higher PSNR values indicate better reconstruction fidelity. \textit{Structural Similarity Index Measure (SSIM)} evaluates the perceived similarity between original and reconstructed signals by considering luminance, contrast, and structure, with SSIM values ranging from 0 to 1, where 1 represents perfect similarity.

\textit{Fréchet Inception Distance (FID)} quantifies the distributional similarity between original and reconstructed feature representations. Lower FID scores correspond to higher fidelity in reconstructed data distributions. The \textit{Correlation Coefficient (CC)} measures the linear correlation between ground truth and reconstructed signals as $ \mathrm{CC} = \frac{\mathrm{cov}(\mathbf{x}, \hat{\mathbf{x}})}{\sigma_x \sigma_{\hat{x}}} $, where $\mathrm{cov}$ denotes covariance, and $\sigma_x$, $\sigma_{\hat{x}}$ are the standard deviations of the true and reconstructed signals, respectively.

All metrics are reported using model weights corresponding to the lowest validation loss observed during training.

$\mathcal{D}$

$\mathcal{D}_{Un-lab}$
\section{Experiments and Results}  \label{sec:results}

In this section, we first present the results of the self-supervised training of the FHRFormer, as described in Section \ref{sec:FHRFormer}.  The best-performing models are then applied in two key tasks: i) \emph{Inpainting}, reconstructing missing segments of FHR signals, and ii) \emph{Forecasting}, predicting future FHR values to enable real-time assessment and early warning in wearable systems. 
\subsection {Self-supervised FHR encoder-based on masking}
The performance of the self-supervised FHRFormer was evaluated across various input patch sizes to determine the optimal configuration for reconstructing masked FHR signals. Table~\ref{tab1} presents a comprehensive comparison of eight performance metrics for experimental values of patch sizes $p_s$ ranging from 30 to 480.

The results indicate a clear trend that smaller patch sizes yield superior reconstruction performance across all metrics. Specifically, the $p_s = $ 30 achieved the best results. We will refer to this best-performing model as \textbf{\textit{Hybrid-30}}, while FHRFormer with $p_s = $ 60 remains the second best and will be denoted as \textbf{\textit{Hybrid-60}}. Overall, the performance indicates that smaller patch sizes allow the FHRFormer to better capture localized patterns and frequency components in the FHR signals, which are critical for accurate reconstruction and applications discussed later. 
Larger patch sizes include more contextual information but might require a deeper architecture as the problem becomes more complex, which again can lead to overfitting.   

Figure~\ref{fig:masking} shows performance using six evaluation metrics across different masking ratios. FHRFormer yields the best results when $\gamma$=15\%, hinting that masking larger portions of the FHR signal actually hinders the reconstruction objective described in Sec.~\ref {sec:objective}. The superior performance at smaller patch sizes and balanced masking ratios aligns with the need to handle short-to-medium-duration signal dropouts caused by sensor displacement or maternal movement, as described in the context of wearable FHR monitors. 

\subsection{Inpainting Application}
Missing data in FHR signals can significantly hinder meaningful analysis and risk classification. Therefore, filling those missing values using a DL-powered algorithm is important to carry out meaningful inpainting. We have adopted FHRFormer to fill the missing values where previously linear interpolation was performed in the preprocessing stage. In other words, we are using missing ranges as masked versions and trying to reconstruct the entire FHR signal and only replace values at the masked indices. 

Figure~\ref{fig:inpainting} shows the inpainting application (in line with the potential use case shown in Figure~\ref{fig:main}), where the first row contains the original FHR acquired from the Moyo device and a binary mask representing noise-removed and missing indices. The second row shows how linear lines connecting missing ranges lose frequency harmonics of the physiological signal. The inpainted versions of a random FHR signal from the test set created by the best FHRFormer (Hybrid-30) and the second-best FHRFormer (Hybrid-60) show meaningful variation in those missing indices. This highlights an interesting aspect of how an intelligent inpainting feature can partially compensate for missing values caused by sensor displacement in the wearable devices. 

\subsection{Forecasting Application}
Forecasting the FHR signal based on recent trends can serve as an intelligent feature in wearable devices by enabling early alerts for anticipated drops in heart rate, thereby supporting timely clinical intervention. Beyond early warning, forecasting helps maintain signal continuity during brief interruptions, enhances the robustness of AI-driven risk assessment, and supports proactive clinical decision-making. We developed a forecasting module using the trained weights from FHRFormer, where the model receives a context window of past data and progressively predicts future values in a recursive fashion. Specifically, each predicted segment is fed back into the model as input to forecast the next segment, allowing multi-step forecasting beyond the initial context window. Through experimentation, we found that a 30-minute (3,600 timesteps) context window and a 15-second (30 timesteps) forecasting window yielded the most reliable results. While we employ progressive forecasting during evaluation, a sliding window inference strategy could be adopted in a deployed system to enable continuous, real-time prediction as new data become available.


Figure~\ref{fig:forecasting} illustrates the forecasting performance on a representative FHR signal from the test set, comparing the best-performing model variant (Hybrid-30) and the second-best (Hybrid-60). The gray region represents the fixed 3600-timestep context window used for prediction. The blue curve denotes the ground truth FHR signal, while the colored forecast includes the model prediction with a 95\% confidence interval. For a visual baseline comparison, we evaluated TimeGPT~\cite{garza2023timegpt}, a foundation model for time-series forecasting. As shown, FHRFormer effectively tracks temporal variations and produces accurate predictions that closely follow the true signal, demonstrating its potential for FHR forecasting applications.

\section{Conclusion and Future work}  \label{sec:conclusion}

In this paper, our contributions are threefold. 

First, we introduced a self-supervised, transformer-based architecture for FHR encoding and reconstruction, the {\bf FHRFormer}, which incorporates a frequency-aware loss function to preserve fidelity in both temporal and spectral domains. This capability is particularly important for reconstructing variable-duration signal gaps. The model was trained on a large dataset comprising 4486 FHR signals collected using the Moyo device, using a self-supervised masking strategy.

Second, we applied FHRFormer for inpainting real missing segments in FHR signals. The reconstructed segments preserve both spatial and spectral characteristics consistent with the surrounding signal, improving continuity and quality for downstream analysis.

Third, we proposed a forecasting approach using FHRFormer to predict future FHR values. This strategy enables proactive signal assessment and has potential for real-time monitoring and early warning in clinical settings.

Future work includes validation of the proposed methods in clinical settings, and potential extension to a multimodal framework that incorporates additional clinical parameters or maternal measurements.  

\section*{Ethical Approval}
The Safer Births 2  project was ethically approved before implementation by the National Institute for Medical Research (NIMR) in Tanzania ( NIMR/HQ/R.8a/Vol. IX/3852) and the Regional Committee for Medical and Health Research Ethics (REK) in Norway, reference 172126. 

Conflict of Interest: K.E. holds a 20\% affiliation with Laerdal Medical; however, the research was conducted independently of this affiliation. The remaining authors declare that they have no conflicts of interest.

\section*{Acknowledgment}
The Laerdal Foundation, and the Research Council of Norway, through the Global Health and Vaccination Program (GLOBVAC - project number 228203), funded the research infrastructure, data collection and management of the Safer Births 2 research project. Idella Fondation funded the post.doc grant of N.K. 
The authors would also like to thank all the mothers and midwives who contributed to the data collection.

\bibliographystyle{IEEEtran}
\bibliography{main}

\end{document}